%% file: main.tex
\documentclass[10pt,twocolumn,letterpaper]{article}

\usepackage[pagenumbers]{cvpr} % To force page numbers, e.g. for an arXiv version

\usepackage{graphicx}
\usepackage{amsmath}
\usepackage{amssymb}
\usepackage{booktabs}
\usepackage{makecell}
\usepackage{multirow}

\usepackage[pagebackref,breaklinks,colorlinks]{hyperref}

\DeclareMathSymbol{\mhyphen}{\mathord}{AMSa}{"39}

\usepackage[capitalize]{cleveref}
\crefname{section}{Sec.}{Secs.}
\Crefname{section}{Section}{Sections}
\Crefname{table}{Table}{Tables}
\crefname{table}{Tab.}{Tabs.}

\usepackage[dvipsnames]{xcolor}
\def\cvprPaperID{3773} % *** Enter the CVPR Paper ID here
\def\confName{CVPR}
\def\confYear{2023}

\begin{document}

\title{CLIP the Gap: A Single Domain Generalization Approach for Object Detection}

\author{Vidit Vidit$^1$ \  Martin Engilberge$^1$ \  Mathieu Salzmann$^{1,2}$ \\
CVLab, EPFL$^1$, ClearSpace SA$^2$\\
{\tt\small {firstname.lastname}@epfl.ch}
}
\maketitle

\input{tex/defs}

\input{tex/abstract}
\input{tex/introduction}
\input{tex/related_work}
\input{tex/method}

\input{tex/experiments}
\input{tex/limit_societal_impact}

\input{tex/conclusion}

%%%%%%%%% REFERENCES
{\small
\bibliographystyle{ieee_fullname}
\bibliography{bibliography}
}

\end{document}

%% file: tex/defs.tex
% !TEX root = ../main.tex
% !TEX spellcheck = en-US

\newif\ifdraft
\drafttrue

\newcommand{\bt}{\mathbf{t}}
\newcommand{\bM}{\mathbf{M}}
\newcommand{\bd}{\mathbf{d}}
\newcommand{\bX}{\mathbf{X}}
\newcommand{\bz}{\mathbf{z}}
\newcommand{\bG}{\mathbf{G}}
\newcommand{\bB}{\mathbf{B}}
\newcommand{\bS}{\mathbf{S}}
\newcommand{\bH}{\mathbf{H}}
\newcommand{\bR}{\mathbf{R}}
\newcommand{\bA}{\mathbf{A}}
\newcommand{\bdelta}{\boldsymbol{\delta}}

\newcommand{\addparagraphup}{\vspace*{0.01in}}

% corrections
\newcommand{\ms}[1]{\ifdraft {\color{green}{#1}} \else {#1}\fi}
\newcommand{\vdt}[1]{\ifdraft {\color{blue}{#1}} \else {#1}\fi}
\newcommand{\me}[1]{\ifdraft {\color{cyan}{#1}} \else {#1}\fi}

% remarks
\newcommand{\MS}[1]{\ifdraft {\color{green}{\textbf{MS: #1}}}\else {}\fi}
\newcommand{\VDT}[1]{\ifdraft {\color{red}{\textbf{VDT: #1}}}\else {}\fi}
\newcommand{\ME}[1]{\ifdraft {\color{cyan}{\textbf{ME: #1}}} \else {}\fi}
\newcommand{\chng}[1]{\ifdraft {\color{red}{#1}}\else {}\fi}

% Citations
\newcommand{\comment}[1]{}
\newcommand{\teaserfig}[1]{./images/#1}

% colors

% \iffalse
\iftrue % use space-saving macro

\newcommand{\cutsectionup}{\vspace*{0in}}
\newcommand{\cutsectiondown}{\vspace*{-0.05in}}

\newcommand{\cutsubsectionup}{\vspace*{-0.05in}}
\newcommand{\cutsubsectiondown}{\vspace*{-0.04in}}

\newcommand{\cutsubsubsectionup}{\vspace*{-0.05in}}
\newcommand{\cutsubsubsectiondown}{\vspace*{-0.05in}}

\newcommand{\cutparagraphup}{\vspace*{-0in}}
\newcommand{\cutparagraphdown}{\vspace*{-0in}}

\newcommand{\cuthalfcaptionup}{\vspace*{-0.15in}}
\newcommand{\cuthalfcaptiondown}{\vspace*{-0.25in}}

\newcommand{\cutcaptionup}{\vspace*{-0.1in}}
\newcommand{\cutcaptiondown}{\vspace*{-0.15in}}

\newcommand{\cuthalftablecaptionup}{\vspace*{-0in}}
\newcommand{\cuthalftablecaptiondown}{\vspace*{-0.1in}}

\newcommand{\cuttablecaptionup}{\vspace*{-0in}}
\newcommand{\cuttablecaptiondown}{\vspace*{-0.1in}}

\newcommand{\cutequationup}{\vspace*{-0.05in}}
\newcommand{\cutequationdown}{\vspace*{-0.01in}}

\newcommand{\cuttableup}{\vspace*{-0.2in}}
\newcommand{\cuttabledown}{\vspace*{-0.3in}}

\newcommand{\cutabstractup}{\vspace*{-0.12in}}
\newcommand{\cutabstractdown}{\vspace*{-0.3in}}

\newcommand{\cutalgorithmup}{\vspace*{-0in}}
\newcommand{\cutalgorithmdown}{\vspace*{-0.1in}}

\newcommand{\cut}{{\vspace*{-0.02in}}}
\newcommand{\cutmore}{{\vspace*{-0.06in}}}
\newcommand{\negcut}{}
\else % do not use space-saving macro
\newcommand{\cutsectionup}{}
\newcommand{\cutsectiondown}{}

\newcommand{\cutsubsectionup}{}
\newcommand{\cutsubsectiondown}{}

\newcommand{\cutsubsubsectionup}{}
\newcommand{\cutsubsubsectiondown}{}

\newcommand{\cutparagraphup}{}
\newcommand{\cutparagraphdown}{}

\newcommand{\cuthalfcaptionup}{}
\newcommand{\cuthalfcaptiondown}{}

\newcommand{\cutcaptionup}{}
\newcommand{\cutcaptiondown}{}

\newcommand{\cutequationup}{}
\newcommand{\cutequationdown}{}

\newcommand{\cuttableup}{}
\newcommand{\cuttabledown}{}

\newcommand{\cutabstractup}{}
\newcommand{\cutabstractdown}{}

\newcommand{\cutalgorithmup}{}
\newcommand{\cutalgorithmdown}{}

\newcommand{\cut}{}
\newcommand{\cutmore}{}
\newcommand{\negcut}{}
\fi

%% file: tex/abstract.tex
\begin{abstract}
    Single Domain Generalization (SDG) tackles the problem of training a model on a single source domain so that it generalizes to any unseen target domain. While this has been well studied for image classification, the literature on SDG object detection remains almost non-existent. To address the challenges of simultaneously learning robust object localization and representation, we propose to leverage a pre-trained vision-language model to introduce semantic domain concepts via textual prompts. We achieve this via a semantic augmentation strategy acting on the features extracted by the detector backbone, as well as a text-based classification loss. 
    Our experiments evidence the benefits of our approach, outperforming  by 10\% the only existing SDG object detection method, Single-DGOD~\cite{wu2022single}, on their own diverse weather-driving benchmark.
%    in the detection task further increases this complexity. Our approach relies on the vision language models which have generalizable semantic embeddings and leverage their language grounding to introduce semantic domain concepts via textual prompts. We outperform Single-DGOD~\cite{wu2022single} on their newly proposed diverse weather-driving SDG benchmark by 10\%.
\end{abstract}  

%% file: tex/introduction.tex
\section{Introduction}
As for most machine learning models, the performance of object detectors degrades when the test data distribution deviates from the training data one. Domain adaptation techniques~\cite{chen2018domain,saito2019strong,chen2020harmonizing,shen2019scl,deng2021unbiased,li2022cross} try to alleviate this problem by learning domain invariant features between a source and a known target domain. In practice, however, it is not always possible to obtain target data, even unlabeled, precluding the use of such techniques. Domain generalization tackles this by seeking to learn representations that generalize to any target domain. While early approaches~\cite{balaji2018metareg,dou2019domain,li2018learning,li2018domain,wang2019towards,zhao2020domain,li2019episodic} focused on the scenario where multiple source domains are available during training, many recent methods tackle the more challenging, yet more realistic, case of Single Domain Generalization (SDG), aiming to learn to generalize from a single source dataset. While this has been well studied for image classification~\cite{volpi2018generalizing,qiao2020learning,fan2021adversarially,wang2021learning,zhang2022exact}, it remains a nascent topic in object detection. To the best of our knowledge, a single existing approach, Single-DGOD~\cite{wu2022single}, uses disentanglement and self-distillation~\cite{kim2020self} to learn domain-invariant features. 

\input{figures/teaser.tex}

In this paper, we introduce a fundamentally different approach to SDG for object detection. To this end, we build on two observations: (i) Unsupervised/self-supervised pre-training facilitates the transfer of a model to new tasks~\cite{he2020momentum,chen2021exploring,caron2021emerging}; (ii) Exploiting language supervision to train vision models allows them to generalize more easily to new categories and concepts\cite{desai2021virtex,radford2021learning}.
%given a small amount of data from the downstream task.
Inspired by this, we therefore propose to leverage a self-supervised vision-language model, CLIP~\cite{radford2021learning}, to guide the training of an object detector so that it generalizes to unseen target domains. 
%Note that, in contrast to~\cite{gu2021open,rasheed2022bridging}, we do not seek to detect new object categories, but instead introduce text-based domain variations during training.
Since the visual CLIP representation has been jointly learned with the textual one, we transfer text-based domain variations to the image representation during training, thus increasing the diversity of the source data.

%It is common in these methods to initialize with ImageNet~\cite{ILSVRC15} weights trained on the supervised classification task. However, in recent years, models fine-tuned with unsupervised/self-supervised pre-training~\cite{he2020momentum,chen2021exploring,caron2021emerging} weights have shown better transferability than the supervised counterparts. Vision models trained with language supervision~\cite{desai2021virtex,radford2021learning} show greater generalizability to new categories and concepts while needing a smaller amount of data for the downstream task. The study of the effect of such pre-trained models is missing in the context of SDG. In this work, we propose to leverage a vision language model, CLIP~\cite{radford2021learning} to better generalize to the unseen target domains. Contrary to other works~\cite{gu2021open,rasheed2022bridging}, which use them to detect new object categories, we want to introduce new domain concepts while training. Since visual representation in these models is jointly learned with the text, one can introduce a new concept description via text to estimate similar visual representation. Text prompt-based image~\cite{ramesh2021zero,ramesh2022hierarchical,saharia2022photorealistic} and video~\cite{singer2022make,ho2022imagen} generation has been feasible because of rich semantic embeddings of these vision language models.

Specifically, we 
%initialize with CLIP weights and then introduce semantic augmentation with the help of the 
define textual prompts describing potential target domain concepts, such as weather and daytime variations for road scene understanding, and use these prompts to perform semantic augmentations of the images. These augmentations, however, are done in feature space, not in image space, which is facilitated by the joint image-text CLIP latent space.
%as it will require a more detailed and diverse text prompt for the latter. 
This is illustrated in~\cref{fig:teaser}, which shows that, even though we did not use any target data for semantic augmentation, the resulting augmented embeddings reflect the distributions of the true image embeddings from different target domains.
%illustrates how the features estimate using text prompts align well with original image features, which we don't have access to during training.

We show the effectiveness of our method on the SDG driving dataset of~\cite{wu2022single}, which reflects a practical scenario where the training (source) images were captured on a clear day whereas the test (target) ones were acquired in rainy, foggy, night, and dusk conditions. %Single-DGOD~\cite{wu2022single} which proposes this dataset also propose a disentanglement and self-distillation~\cite{kim2020self} method to learn domain-invariant features. By contrast, we want to learn robust features to help improve the generalization. 
Our experiments demonstrate the benefits of our approach over the Single-DGOD~\cite{wu2022single} one.

To summarize our contributions, we employ a vision-language model 
%CLIP~\cite{radford2021learning} 
to improve the generalizability of an object detector; during training, we introduce domain concepts via text-prompts to augment the diversity of the learned image features and make them more robust to an unseen target domain. This enables us to achieve state-of-the-art results on the diverse weather SDG driving benchmark of~\cite{wu2022single}.

%% file: figures/teaser.tex
\begin{figure}
    \includegraphics[width=\linewidth]{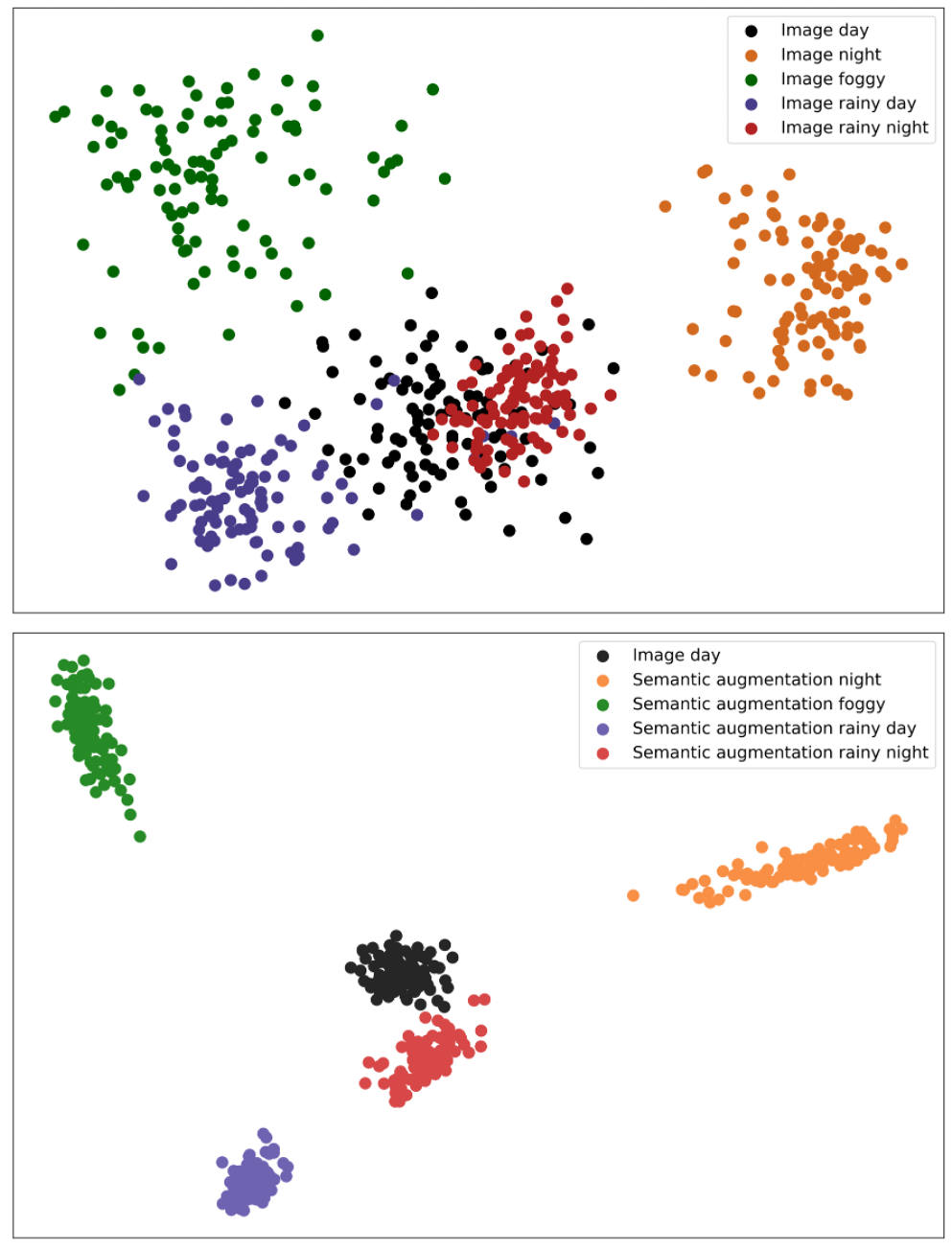}
    \caption{\textbf{Semantic Augmentation:} We compare the PCA projections of CLIP~\cite{radford2021learning} image embeddings obtained in two different manners: (Top) The embeddings were directly obtained from the real images from 5 domains corresponding to different weather conditions. (Bottom) The embeddings were obtained from the \emph{day} images only and modified with our semantic augmentation strategy based on text prompts to reflect the other 4 domains. Note that the relative positions of the clusters in the bottom plot resembles that of the top one, showing that our augmentations let us generalize to different target domains. The principal components used are the same for both the figures.
    }
    %We visualize the gap between domains in the CLIP~\cite{radford2021learning} embeddings space using PCA projection. The top image depicts the clip representation of images coming from 5 different weather conditions.
    %The bottom one depicts the clip representation of images from the clear day set extended using our semantic augmentation method. By using only text prompts, our method is able to mimic the shift between domains. Note that the relative positioning of the cluster is similar between the top and bottom plots even though the bottom one only uses images from the clear day dataset.}

    % "weather-concept"-I corresponds to image embeddings and "weather-concept"-T depicts translated \emph{day}-I features according to our augmentation strategy. By using only the text prompts, \emph{day}-I image features are translated to different concepts like \emph{foggy/rainy}. The relative position of translated features is similar to original image ones (relative to \emph{day}-I) as the CLIP~\cite{radford2021learning} learns a joint image and text embeddings.}
    \label{fig:teaser}
\end{figure}

%% file: tex/related_work.tex
\section{Related Work}

\paragraph{Domain Adaptation for Object Detection.} Domain adaptation methods seek to align the source domain distribution to a particular target domain. To bridge the global and instance-level domain gaps, \cite{chen2018domain,saito2019strong,chen2020harmonizing,shen2019scl} learn feature alignment via~\cite{ganin2016domain} adversarial training; \cite{zhu2019adapting} and  \cite{vs2021mega} utilize category-level centroids and attention maps, respectively, to better align instances in the two domains; \cite{deng2021unbiased,li2022cross} generate pseudo-labels in the target domain and use them for target-aware training. Domain adaptation, however, assumes that images from the target domain are available during training. In contrast, domain generalization aims to learn models that generalize to domains that were not seen at all during training. Below, we focus on the domain generalization methods that, as us, use a single source domain to do so.

\paragraph{Single Domain Generalization (SDG).} Several image classification works~\cite{volpi2018generalizing,qiao2020learning,fan2021adversarially,wang2021learning,zhang2022exact} have proposed strategies to improve the performance on \emph{unseen} domains while training on a single source domain. In particular, \cite{volpi2018generalizing,qiao2020learning,wang2021learning} introduce data augmentation strategies where diverse input images are generated via adversarial training; \cite{fan2021adversarially,zhang2022exact} propose normalization techniques to adapt the feature distribution to unseen domains. While SDG has been reasonably well studied for image classification, the case of object detection remains largely unexplored, and poses additional challenges related to the need to further localize the objects of interest.
%and learning robust foreground object representations pose an additional challenge in detection scenarios. 
This was recently tackled by Single-DGOD~\cite{wu2022single} with an approach relying on learning domain-specific and domain-invariant features. Specifically, this was achieved by exploiting contrastive learning to disentangle the features and self-distillation~\cite{kim2020self} to further improve the network's generalizability.
%propose a novel approach and benchmark dataset for SDG under object detection setting.
Here, we introduce a fundamentally different approach that leverages the CLIP~\cite{radford2021learning} pre-trained model and semantically augments the data using textual prompts. As will be shown by our results, our method outperforms the state-of-the-art Single-DGOD~\cite{wu2022single}.

\paragraph{Vision-Language Models.}
Jointly learning a representation of images and text has been studied in many works~\cite{frome2013devise,kiros2014unifying,engilberge2018finding,faghri2017vse++,li2019unicoder,zhang2020contrastive,desai2021virtex,radford2021learning}. They use image-text pairs to train visual-semantic embeddings which can be used not only for image classification, captioning or retrieval but also for zero-shot prediction on unseen labels. VirTex~\cite{desai2021virtex} relies on image-caption-based pre-training to learn a rich visual embedding from a small amount of data. CLIP~\cite{radford2021learning} proposes a scalable contrastive pre-training method for joint text and image feature learning. CLIP leverages a corpus of 400 million image-text pairs and a large language model~\cite{radford2019language} to learn a joint embedding space, which was shown to have superior zero-shot learning ability on classification tasks. The image-text-based training is also useful for Open Vocabulary Detection (OVD)~\cite{zareian2021open}, where the objects are detected using arbitrary textual descriptions. To address this task, \cite{zareian2021open} train their own visual-semantic representation, whereas ~\cite{gu2021open,rasheed2022bridging} employ CLIP embeddings. Recently, \cite{li2022grounded,zhang2022glipv2} introduced a phrase-grounding-based pre-training for better OVD and zero-shot object detection. In contrast to these works, whose objective is to generalize to novel \emph{categories or objects}, we seek to generalize to new \emph{domains} depicting the same object categories as the source one. 
%The domain change between the dataset can be attributed to the global illumination or appearance change.

%% file: tex/method.tex
\section{Method}
\input{figures/arch.tex}    

Let us now introduce our approach to exploiting a vision-language model for single-domain generalization in object detection. Below, we first present our semantic augmentation strategy aiming to facilitate generalization to new domains. We then describe the architecture and training strategy for our object detector.

\subsection{Semantic Augmentation}
\label{sec:augmentation}
In SDG, we have access to images from only a single domain. To enable generalization, we seek to learn object representations that are robust to domain shifts. Here, we do so by introducing such shifts while training the model on the source data. Specifically, we exploit CLIP's joint representation to  estimate shifts in the visual domain using textual prompts, as illustrated in~\cref{fig:teaser}. This corresponds to the optimization step shown in the left portion of~\cref{fig:arch}.

Formally, let $\mathcal{T}$ denote CLIP's text encoder and $\mathcal{V}$ its image one. For reasons that will become clear later, we further split $\mathcal{V}$ into a feature extractor $\mathcal{V}^{a}$ and a projector to the embedding space $\mathcal{V}^{b}$. The CLIP~\cite{radford2021learning} model is trained to bring image features closer to their textual captions. In essence, this means that, for an image $\mathcal{I}$ and a corresponding prompt $p$, it seeks to minimize the distance between $\mathcal{V}^b(\mathcal{V}^a(I))$ and $\mathcal{T}(p)$.

A useful property of the text embedding space is that algebraic operations can be used to estimate semantically related concepts.  Word2Vec~\cite{mikolov2013efficient} had demonstrated such a learned relationship (e.g. \emph{king}-\emph{man}+\emph{woman} approaches the word representation of \emph{queen}). Such a relationship exists with CLIP embeddings as well~\cite{ramesh2022hierarchical}.

To exploit this for SDG, we define a generic textual prompt ${p}^s$ related to the source domain, such as \texttt{An image taken during the day}, and a set of prompts $\mathcal{P}^t=\{p_j^t\}_1^M$ encompassing variations that can be expected to occur in different target domains, e.g, describing different weather conditions or times of the day.
Our objective then is to define augmentations $\{\mathcal{A}_j\}$ of the features extracted from a source image such that the shift incurred by $\mathcal{A}_j$ corresponds to the semantic difference
%model the shift in feature space corresponding to the difference 
between $p^s$ and $p^t_j$. 
%That is, for each prompt in $\mathcal{P}^t$,  we search for an augmentation $\mathcal{A}_j \in \mathbb{R}^{H\times W\times C}$ of the image features extracted by $\mathcal{V}^{a}$ such that the difference between the embeddings extracted by $\mathcal{V}^{b}$ with and without augmentation is similar to that between the embeddings obtained from $p^t_j$ and $p^s$.

To achieve this, we first compute the embeddings $q^s=\mathcal{T}(p^s)$ and $q^t_j = \mathcal{T}(p^t_j)$ of the textual prompt. We then take multiple random crops from a source image. For each such crop $\mathcal{I}_{crop}$, we create a target image embedding
\begin{equation}
    z^{\ast}_j = {z}+\frac{q^t_j-q^s}{\|q^t_j-q^s\|_2 }\;,
\end{equation}
where ${z} =\mathcal{V}(\mathcal{I}_{crop})$. We then search for an augmentation $\mathcal{A}_j \in \mathbb{R}^{H\times W\times C}$ such that
\begin{align}
    \bar{z}_j &= \mathcal{V}^{b}(\mathcal{V}^{a}(\mathcal{I}_{crop})+\mathcal{A}_j) \label{eq:feattsl}
\end{align}
is as similar as possible to $z^{\ast}_j$, which we measure with the cosine similarity.
\input{figures/dataset.tex}
Ultimately, we estimate the augmentations $\{\mathcal{A}_j\}_1^M$ through an optimization process using only source domain images. Specifically, we minimize the loss function
\begin{equation}
    \mathcal{L}_{opt}=\sum_{\mathcal{I}_{crop}}\sum_{j}{\mathcal{D}(z^{\ast}_j ,\bar{z}_j) + \|\bar{z}_j-z\|_1}\;, \label{eq:opt}
\end{equation}
where
\begin{equation}
    \mathcal{D}(a,b) = 1-\frac{a-b}{\|a-b\|_2} 
\end{equation}
is the cosine distance. The loss also includes an $l_1$ regularizer that prevents the embeddings from deviating too far from their initial values, so as to preserve the image content. 

As the objective is to estimate the meaningful feature augmentation while preserving the original CLIP pre-training, we keep the image crop size the same as the original CLIP training.  Note that the optimization of the augmentations is done once in an offline stage, and we then use the resulting augmentations to train our detector.

\subsection{Architecture}

%The CLIP~\cite{radford2021learning} model is trained to bring image features closer to their textual captions. While a visual encoder, $\mathcal{V}$, provides an appropriate representation of an image and a text encoder, $\mathcal{T}$, projects the prompt to a joint space in $\mathbb{R}^{D_{clip}}$. The prompts are  tokenized and transformed into word embedding before being given as input to  $\mathcal{T}$. 

%Though, these models learn a broader set of object categories and scene descriptions but lack object localization through their prompts. Hence, 

Let us now describe our detector architecture. As shown in the right portion of~\cref{fig:arch}, it follows a standard FasterRCNN~\cite{ren2016faster} structure but departs from it in two ways. First, to exploit the augmentations optimized as discussed in the previous section, we initialize the blocks before and after the ROI align one with the corresponding $\mathcal{V}^a$ and $\mathcal{V}^b$ modules of the ResNet-based trained CLIP model. Second, to further leverage the vision-language model, we incorporate a text-based classifier in our model's head. Note that, in contrast to OVD~\cite{gu2021open,rasheed2022bridging} where a text-based classifier is used to handle novel categories, we employ it to keep the image features close to the pre-trained joint embedding space.

%and incorporate a text-based classifier. We separate $\mathcal{V}$ as $\mathcal{V}^{a}$, which serves as a feature extractor for FasterRCNN and $\mathcal{V}^{b}$ which takes regions proposed by Region Proposal Network (RPN) as input. $T$ is used for classification.

Specifically, we define textual prompts that represent the individual categories we seek to detect, and extract corresponding embeddings $\mathcal{Q}\in \mathbb{R}^{(K+1)\times D_{clip}}$, for $K$ categories and the background class, using the text encoder $\mathcal{T}$. For a candidate image region $r$ proposed by the Region Proposal Network(RPN)~\cite{ren2016faster}, we then compute the cosine similarities between the text embeddings $\mathcal{Q}$ and the features $\mathcal{F}_r \in \mathbb{R}^{D_{clip}}$ obtained by projection to the embedding space using $\mathcal{V}^{b}$ after ROI-Align~\cite{he2017mask} and the text embeddings $\mathcal{Q}$. These cosine similarities, $sim(\mathcal{F}_r,\mathcal{Q}) \in \mathbb{R}^{K+1}$, act as logits to the softmax based cross-entropy loss
\begin{equation}
    \mathcal{L}_{clip\mhyphen t} = \sum_{r}{\mathcal{L}_{CE}\left(\frac{e^{sim(\mathcal{F}_r,\mathcal{Q}_k)}}{\sum_{k=0}^{K}{e^{sim(\mathcal{F}_r,\mathcal{Q}_k)}}}\right)}\;.
    \label{eq:clip_loss}
\end{equation}
Similarly to~\cite{radford2021learning}, we formulate prompts of the form  \texttt{a photo of a \{category name\}} to obtain our text embeddings.

\subsection{Training with Augmentation}
Following the standard detector training~\cite{ren2016faster}, we use the full image as our input. This subsequently increases the output feature map size of $\mathcal{V}^a$, hence we use average pooling operation and obtain channel-wise augmentations which can work for arbitrary-sized feature maps. The training of our modified object detector with the semantic augmentations is as follows, first, we randomly sample an augmentation $\mathcal{A}_j$ from the full set and collapse its spatial dimension using average pooling. We then add the resulting vector to every element in the feature map extracted by $\mathcal{V}^a$. In practice, we apply augmentations to a batch with a probability $\theta$.

%batch to apply these augmentations. Additionally, we collapse the spatial dimension of $\mathcal{A}$ using \emph{avgpool} operation and add it to the feature maps. 

The detector is then trained with the loss 
\begin{equation}
    \mathcal{L}_{det} = \mathcal{L}_{rpn} + \mathcal{L}_{reg} + \mathcal{L}_{clip\mhyphen t}\;,
\end{equation}
which combines the $\mathcal{L}_{clip\mhyphen t}$ loss of~\cref{eq:clip_loss} with the standard RPN and regression losses~\cite{ren2016faster}. During inference, we use the detector without any augmentation of the feature maps.

%% file: figures/arch.tex
\begin{figure*}
    \centering
    \includegraphics[width=\textwidth]{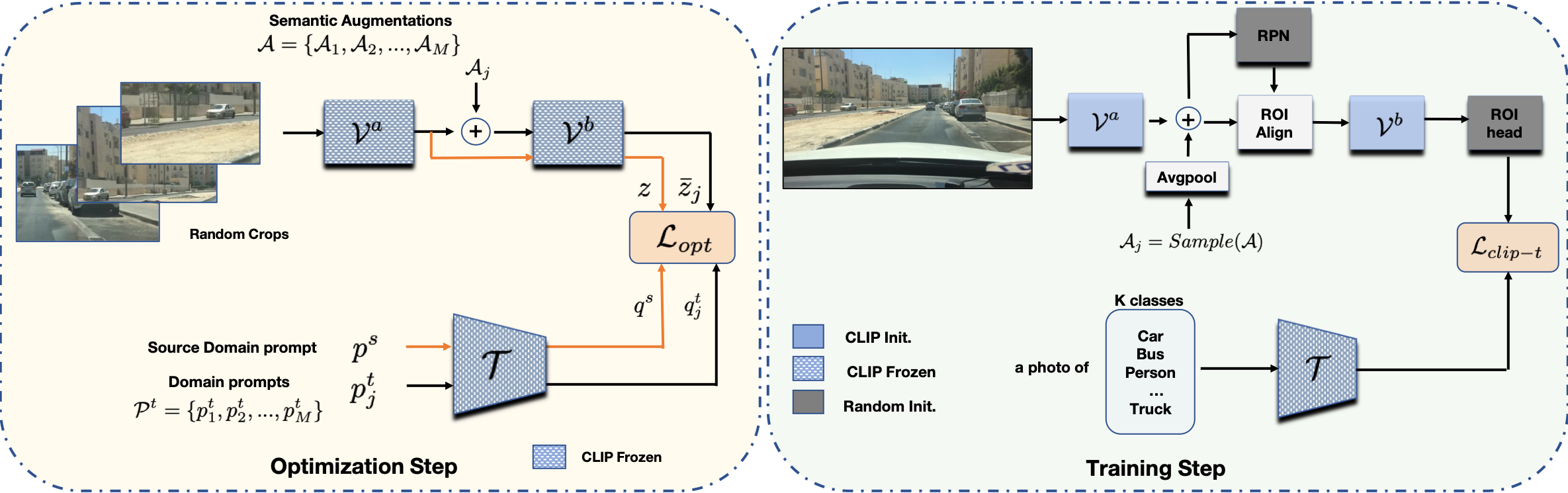}
    \caption{\textbf{Our Approach:} {\bf (Left)} We first estimate a set of semantic augmentations $\mathcal{A}$ using a set of textual domain prompts $\{\mathcal{P}^t,p^s\}$ and source domain images.  The goal of these semantic augmentations is to translate source domain image embeddings to the domain specified by the prompts. We can do this because of the CLIP's joint embedding space and its ability to encode semantic relationships via algebraic operations. $\mathcal{L}_{opt}$ is minimized w.r.t $\mathcal{A}$ over random image crops of the same size as CLIP~\cite{radford2021learning}. {\bf (Right)} The optimized semantic augmentations are used to train our modified detector which minimizes a text-based classification loss $\mathcal{L}_{clip\mhyphen t}$. Here, we train with the full image and add a randomly sampled $\mathcal{A}_j$ after average pooling. This pooling operation allows us to use  $\mathcal{A}$ on extracted feature maps of the arbitrary-sized image. We initialize the detector with the pre-trained CLIP~\cite{radford2021learning} $\mathcal{V}$ and $\mathcal{T}$ encoders to leverage their general representations.}
    \label{fig:arch}
\end{figure*}

%% file: figures/dataset.tex
\begin{figure*}
    \centering
    \includegraphics[width=\linewidth]{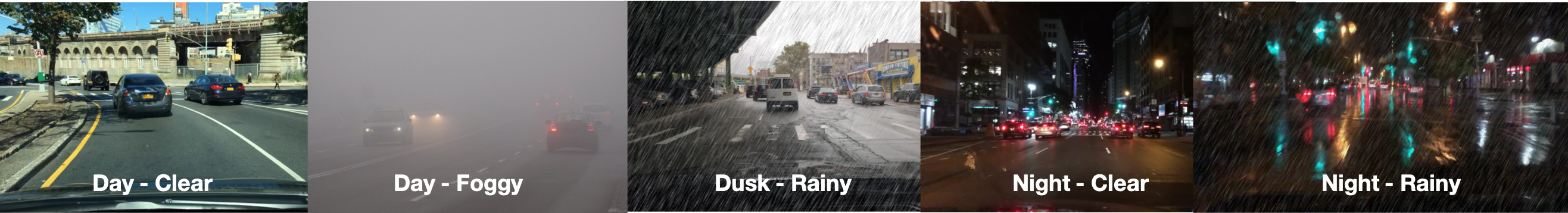}
    \caption{\textbf{Diverse Weather Dataset~\cite{wu2022single}:} Day-Clear acts as our source domain while the other weather condition are our target domains. In these domains, the objects' appearance drastically changes from the Day-Clear scenario. As we do not utilize any target domain images, learning generalizable features on source images is crucial for the SDG task. }
    \label{fig:dataset}
\end{figure*}

%% file: tex/experiments.tex
\section{Experiments}
\subsection{Experimental setup}

\paragraph{Datasets.} 
To evaluate our model, we use the same datasets as~\cite{wu2022single}. They include five sets, each containing images with different weather conditions: daytime sunny, night clear, dusk rainy, night rainy, and daytime foggy.  The images have been selected from three primary datasets, Berkeley Deep Drive 100K (BBD-100K) \cite{yu2020bdd100k}, Cityscapes \cite{cordts2016cityscapes} and Adverse-Weather \cite{hassaballah2020vehicle}. Additionally, rainy images are rendered by \cite{wu2021vector}, and some of the foggy images are synthetically generated from \cite{sakaridis2018semantic}.
Our model is trained on the daytime sunny scenes, consisting of 19,395 training images, the remaining 8,313 daytime sunny images are used for validation and model selection. The four other weather conditions are only used during testing. They consist of 26,158 images of clear night scenes, 3501 images of rainy scenes at dusk, 2494 images of rainy scenes at night, and 3775 images of foggy scenes during daytime. All the datasets contain bounding box annotations for the objects \emph{bus, bike, car, motorbike, person, rider} and \emph{truck}.~\cref{fig:dataset} shows examples from this dataset.

\paragraph{Metric.} In all our experiments, we use the Mean Average Precision (mAP) as our metric. Specifically, following~\cite{wu2022single}, we report the mAP@0.5, which considers a prediction as a true positive if it matches the ground-truth label and has an intersection over union (IOU) score of more than 0.5 with the ground-truth bounding box.
 \input{figures/qual_results.tex}
\input{figures/foggy.tex}

 \subsection{Implementation Details}
 We use the Detectron2~\cite{wu2019detectron2} implementation of FasterRCNN with a ResNet101~\cite{he2016deep} backbone. We initialize the detector with CLIP~\cite{radford2021learning} pre-trained weights, where ResNet convolution blocks 1-3 act as $\mathcal{V}^a$, and block-4 along with the CLIP attention pooling act as $\mathcal{V}^b$. This follows from the standard FasterRCNN implementation with ResNet backbone.
\paragraph{Optimization Step.} As the benchmark dataset evaluates the method on different weather conditions, we curated a list of domain prompts $\mathcal{P}^t$ matching the concept \emph{weather}. To this end, we take all the \emph{hyponyms} of the term \emph{weather} from WordNet~\cite{wordnet} and generate their text embeddings using the CLIP text encoder $\mathcal{T}$. We prune away the words whose cosine similarity with the term \emph{weather} is lower than $0.5$. Additionally, we filter out the words that are not in the top $10$k frequent words in GloVe wordlist~\cite{pennington2014glove}. After combining the synonyms, we get to a list of six words: \emph{snow, fog, cloudy, rain, stormy, sunshine}. We remove \emph{sunshine} as it corresponds to our source domain concept. Furthermore, we consider three times of the day: \emph{day, night, evening}. This lets us generate $M=15$  prompts using the template \texttt{an image taken on a \{weather\} \{time of the day\}}. We use \texttt{an image taken during the day} as the source domain prompt $p^s$. We provide more details in our supplementary material.

To optimize the augmentations with these prompts, we generated random crops from the source images and resized them to $224\times 224$ pixels. The resulting output feature map of $\mathcal{V}^a$ and $\mathcal{A}_j $ are in $\mathbb{R}^{14\times 14\times 1024}$. We initialize $\mathcal{A}_j \; \forall \; 1\geq j\geq M $ with zeros and train it using the Adam~\cite{kingma2014adam} optimizer while keeping the CLIP encoder, $\mathcal{V}$ and $\mathcal{T}$, frozen. Optimization
%Training 
was done for $1000$ iterations with a learning rate of $0.01$.

\paragraph{Detector Training with Augmentation.}
When training the detector, the input image is resized to $600\times 1067$ and $\mathcal{V}$ and $\mathcal{T}$ are initialized with CLIP pre-trained weights. While $\mathcal{T}$ is kept frozen during the training, the ResNet blocks 3-4 and attention pooling of $\mathcal{V}$, along with the other FasterRCNN learnable blocks, are trained with Stochastic Gradient Descent (SGD) for 100k iterations. We train with a learning rate of $1e^{-3}$, scaled down by a factor of $0.1$ after 40k iterations. We use a batch size of $4$ and apply $\mathcal{A}_j$ to the features with probability $\theta=0.5$. We also use random horizontal flipping augmentation as in Single-DGOD~\cite{wu2022single}. $D_{clip}$ is set to 512 as in~\cite{radford2021learning} and background class is initialized by zeros in $\mathcal{Q}$. All of our training was done on a single NVIDIA A100 GPU. Our code will be made public upon acceptance.
 
\subsection{Comparison with the State of the Art}
We compare our method trained with semantic augmentations against the state-of-the-art Single-DGOD~\cite{wu2022single}. Similar to them, we also show comparisons with feature normalization methods, SW~\cite{pan2019switchable}, IBN-Net~\cite{pan2018two}, IterNorm~\cite{huang2019iterative}, and ISW~\cite{choi2021robustnet}. These methods improve network generalization by using better feature normalization. We additionally report the performance of FasterRCNN (FR) initialized with ImageNet pre-trained weights. For the SDG task, we evaluate the generalization performance on unseen target domains, hence we compare the mAP scores on the out-of-domain datasets: day-foggy, night-rainy, dusk-rainy, and night-clear.

\input{tables/sota}
Our approach of combining CLIP pre-training and semantic augmentation outperforms the baselines on all of the target domains.~\cref{tab:sota} shows a consistent improvement in all domains with close to $15$\% improvement on day-foggy and dusk-rainy compared to Single-DGOD. In the challenging scenario with Night conditions, we improve by $12.6$\% on night-rainy while being comparable with Single-DGOD on night-clear. On the source domain, both our method and Single-DGOD are better than the FR baseline. However, while Single-DGOD gains improvement at the cost of losing out for domain generalization, we improve on both the source and target domains. The failure of feature normalization baselines suggests a large domain gap between the source and target domains.~\cref{fig:qual_result} and~\cref{fig:foggy} provide a qualitative results on different weather-datasets. 

In the remainder of this section, we discuss the per-class results on the individual target domains.

\paragraph{Daytime Clear to Day Foggy.}
\input{tables/class_results_day_foggy}

The object appearance drastically changes in the foggy images compared to the day-clear scenario. As shown in \cref{tab:class_day_foggy}, our method brings in a large improvement for the \emph{car}, \emph{person}, and \emph{bike} categories, while still being consistently better than Single-DGOD and FR on the others.

\input{tables/class_results_dusk_rainy}

\paragraph{Daytime Clear to Dusk Rainy.}

Dusk Rainy scenes reflect a low light condition and along with the rainy pattern. The image distribution is thus further away from the daytime clear images. As shown in~\cref{tab:class_dusk_rainy}, our method improves the AP of each class, with the biggest improvement in the \emph{car} and \emph{person} categories. Since we leverage CLIP pre-training and bring in concepts such as rain/cloudy/stormy and evening/night hours through our semantic augmentation, the learnt detector generalizes better.

\input{tables/class_results_night_clear}

\input{tables/class_results_night_rainy}

  \input{tables/ablation}
\paragraph{Daytime Clear to Night Clear.}
The Night Clear dataset shows a challenging night driving scene under severe low-light conditions. In~\cref{tab:class_night_clear}, we show that while being comparable to Single-DGOD, we bring in a larger improvement in the \emph{car} and \emph{person} categories. Night scenes are particularly challenging as the low light condition leads to more confusion among visually closer categories such as \emph{bus} and \emph{truck}. 

\paragraph{Daytime Clear to Night Rainy.}
This is the most challenging scenario where dark night conditions are exacerbated by patterns occurring due to rain. \cref{tab:class_night_rainy} shows consistent improvement by our approach for most of the classes. The \emph{car} class sees the biggest improvement with an increase in AP of more than 22\% compared to Single-DGOD. The lower performance of the class \emph{rider} can be attributed to an increase in the confusion between the visually similar \emph{person} and \emph{rider} classes under adverse conditions.

\subsection{Ablation Study}
To understand how each element of the proposed method contributes to the overall performance, we conduct an ablation study. We test five individual components of our model. Specifically, we remove semantic augmentation, replace CLIP attention pooling in $\mathcal{V}^b$ with average pooling, replace $\mathcal{L}_{ clip\mhyphen t}$ with the FasterRCNN classification loss, and change the weight initialization from the CLIP model to an ImageNet classification model. Removing those five components turns our model back into the standard FasterRCNN. The ablation study results are provided in \cref{tab:abla} and discussed below.
  
\paragraph{CLIP initialization.} When the FasterRCNN backbone $\mathcal{V}$ is initialized with CLIP pre-trained weights, the model performance consistently increases both in the in-domain and out-of-domain scenarios, as shown in the second row of \cref{tab:abla}. This setting itself already outperforms Single-DGOD (penultimate row of \cref{tab:sota}). This goes to show that, for the generalization task, model weight initialization plays a crucial role. We further improve this performance with semantic augmentations.

% Our method with semantic augmentation further improves the performance by a greater margin on dusk-rainy, night-rainy, day-foggy and remains comparable on the night-clear dataset.

\paragraph{Attention pooling and $\mathcal{L}_{clip\mhyphen t}$.}
Next we test the impact of the text-embedding-based loss $\mathcal{L}_{clip\mhyphen t}$ for classification.
As visible in the third row of \cref{tab:abla}, when combined with CLIP initialization, it improves the generalization performance for the rainy scenarios, but degrades it for the other ones. Replacing average pooling in $\mathcal{V}^b$ with CLIP attention pooling helps to mitigate the detrimental effect of $\mathcal{L}_{clip\mhyphen t}$ and exhibits consistent improvement on all datasets.

\paragraph{Semantic augmentation.} Finally, adding semantic augmentation gives us the best results, as shown in the last row of \cref{tab:abla}. 
%The semantic augmentation build on top of the text embedding-based classification. 
Exposing the visual encoder $\mathcal{V}$ to targeted semantic augmentations helps the overall model to better generalize when exposed to new domains sharing similarity with the augmentations.

%   \paragraph{Relevance of Model Components} ~\cref{tab:abla} contains the ablation study results. As previously mentioned CLIP initialization already improves over the Single-DGOD on all the datasets except night-rainy. We leverage this initialization to get further consistent improvements. We see that $\mathcal{L}_{clip\mhyphen t} $ improves the generalization when we include the attention pooling module of CLIP and train with its pre-trained weights. Since the attention pooling module projects the visual features into the joint embedding of text and image, using text embedding-based loss $\mathcal{L}_{clip\mhyphen t} $ is meaningful. After combining with the semantic augmentation, we improve on generalization further.

 \subsection{Additional Analyses}
 \paragraph{Study of semantic augmentation.} Our proposed method involves translating feature maps by semantic augmentations learned using plausible domain prompts. To further study the utility of our approach, we replace the augmentation strategy in our training pipeline with (a) \textbf{no-aug}: no augmentation; (b) \textbf{random}: $\mathcal{A}$ is initialized with a normal distribution; (c) \textbf{clip-random}: we define $\mathcal{P}^t$ with concepts that are not specific to \emph{weather}. We generate prompts with a template \texttt{an image of \{word\}}, where the words are \emph{desert}, \emph{ocean}, \emph{forest}, and \emph{mountain}. \cref{tab:seg-aug} illustrates the importance of the semantics in our augmentation strategy. The \textbf{random} augmentation performs worse than the \textbf{no-aug} strategy. \textbf{clip-random} is comparable to \textbf{no-aug} and doesn't show any consistent trend but is mostly better than \textbf{random}. Our semantic augmentation strategy provides a consistent improvement over  \textbf{no-aug} because the translations are performed with prompts from the relevant \emph{weather} concept.
 \input{tables/different_aug.tex}

%% file: figures/qual_results.tex
\begin{figure*}
\centering
    \includegraphics[width=\textwidth]{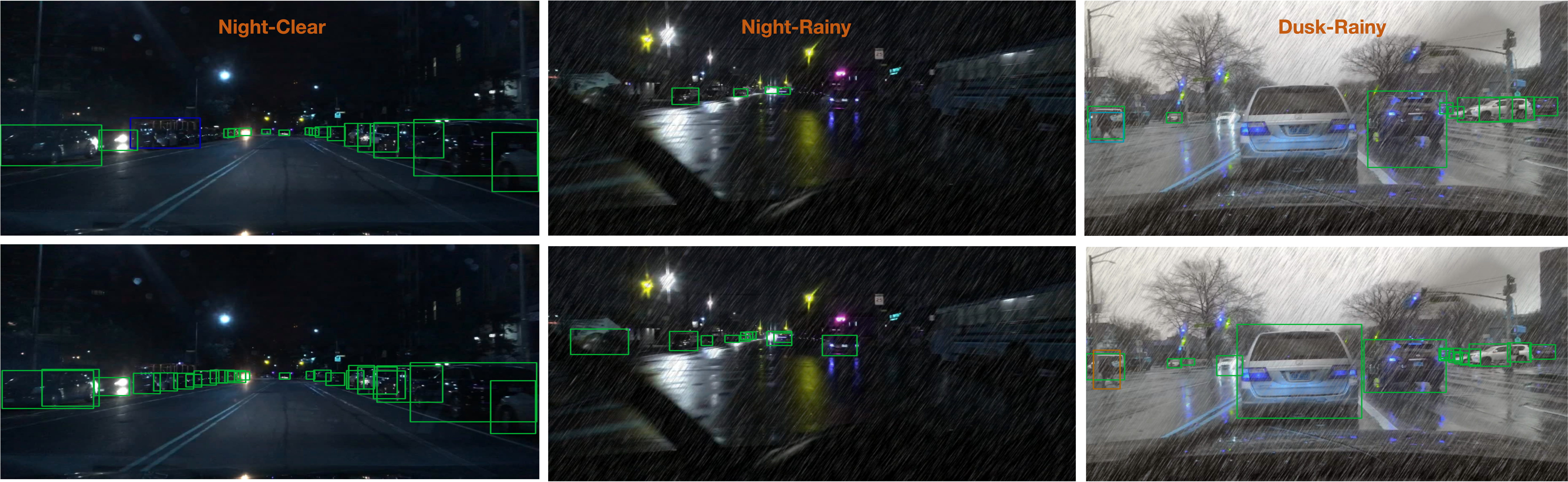}
    \caption{\textbf{Qualitative Results.} We visualize the predictions of the detectors trained only with day-clear images.  (Top) FasterRCNN~\cite{ren2016faster} predictions. (Bottom) The predictions with our approach. Night-Clear and Night-Rainy contain scenes that are taken under low light conditions. Due to this, the appearance of the object is obscure and deviates from the daytime case. FasterRCNN fails to detect most of the objects. As shown in the Night-Clear, it misclassifies a \textcolor{green}{car} to \textcolor{blue}{bus}. By contrast, we can still detect \textcolor{green}{car} under such a big shift. For Dusk-Rainy scenes, the rain pattern on the windscreen and the wet ground causes an appearance shift. As shown FasterRCNN fails to detect several \textcolor{green}{cars} and misclassifies \textcolor{orange}{person} on the bottom-left. }
    \label{fig:qual_result}
\end{figure*}

%% file: figures/foggy.tex
\begin{figure}
    \centering
    \includegraphics[width=\linewidth]{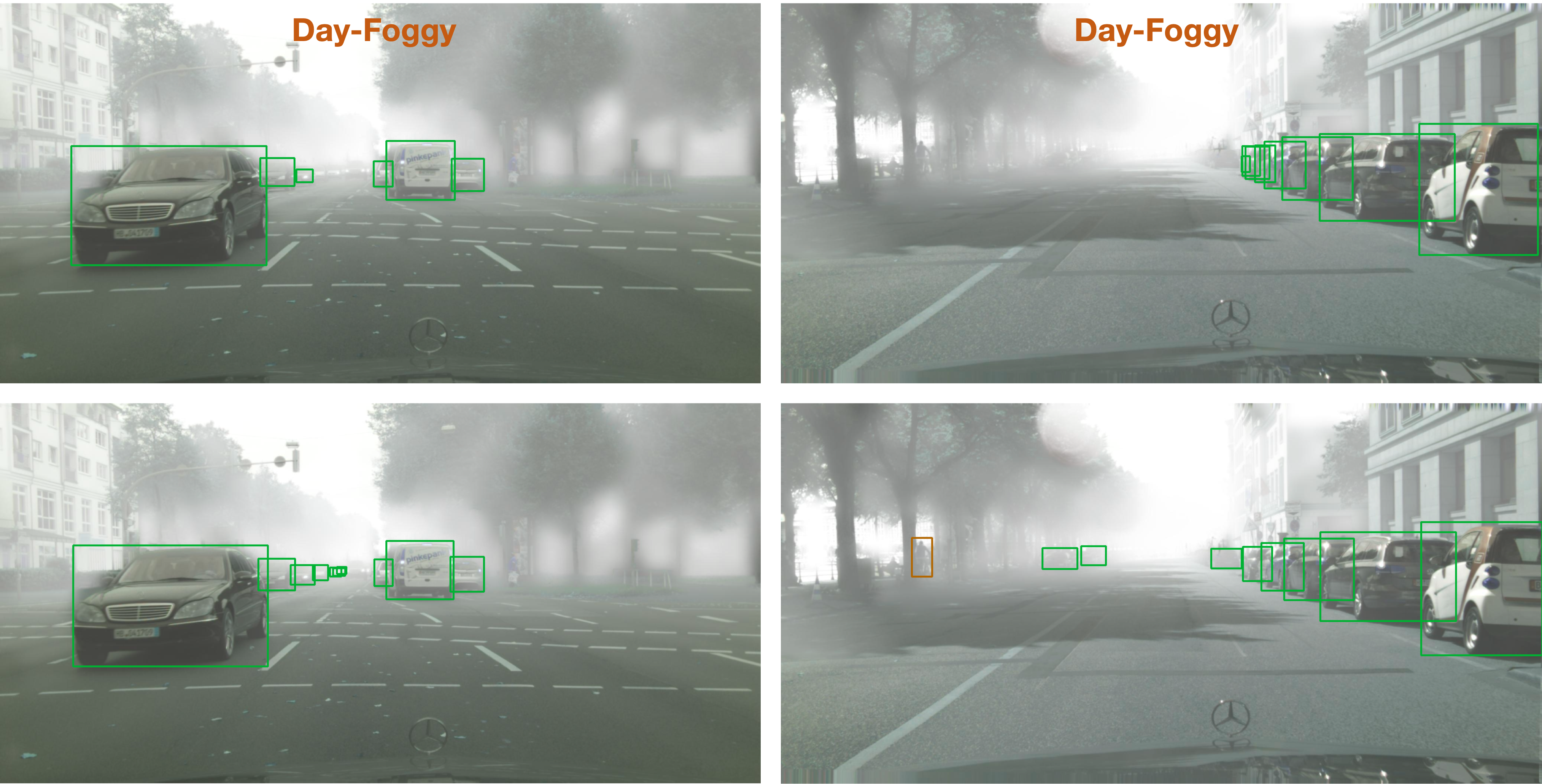}
    \caption{\textbf{Qualitative Results.} In the foggy scenes, the objects further away w.r.t the camera are more obscure than the near ones. Due to this FasterRCNN (Top) struggles to detect them. \textcolor{green}{car} and \textcolor{orange}{person} missed by FasterRCNN are successfully recovered by our approach (Bottom).}
    \label{fig:foggy}
\end{figure}

%% file: tables/sota.tex
\begin{table}[t]
 \setlength{\tabcolsep}{4pt}
    \begin{center}
    % \resizebox{\linewidth}{!}{
    \renewcommand{\arraystretch}{1.5}
    \begin{tabular}{r | c | c c c c } \toprule
    \multicolumn{1}{c}{} & \multicolumn{5}{c}{mAP} \\ \cmidrule(lr{.75em}){2-6}
    %  \multicolumn{1}{c}{} &  &  \multicolumn{4}{c}{Out-Of-Domain} \\ 
     \multicolumn{1}{c}{Method} & \makecell{Day \\  Clear}  & \makecell{Night \\ Clear} & \makecell{Dusk \\ Rainy} & \makecell{Night \\ Rainy} & \makecell{Day \\ Foggy} \\ \midrule
    FR~\cite{ren2016faster} & 48.1 & 34.4 & 26.0 & 12.4 & 32.0 \\
    SW \cite{pan2019switchable} & 50.6 & 33.4 & 26.3 & 13.7 & 30.8 \\
    IBN-Net \cite{pan2018two} & 49.7 & 32.1 & 26.1 & 14.3 & 29.6 \\
    IterNorm \cite{huang2019iterative} & 43.9 & 29.6 & 22.8 & 12.6 & 28.4 \\
    ISW \cite{choi2021robustnet} & 51.3 & 33.2 & 25.9 & 14.1 & 31.8 \\
    S-DGOD \cite{wu2022single} & \textbf{56.1} & 36.6 & 28.2 & 16.6 & 33.5 \\
    \hline
    Ours & 51.3 & \textbf{36.9} & \textbf{32.3} & \textbf{18.7} & \textbf{38.5} \\
    \bottomrule
    \end{tabular}
    % }
    \end{center}
    \caption{\textbf{Single domain generalization results.} We show consistent improvements across all the target domains. S-DGOD boosts the source domain results, but at the cost of reduced generalization ability. By contrast, our approach is robust to domain changes. The numbers for S-DGOD, SW, IBN-Net, IterNorm, ISW are taken from \cite{wu2022single}.
    } 
    \label{tab:sota}
    \end{table}

%% file: tables/class_results_day_foggy.tex
\begin{table}[t]
 
    \begin{center}
    \resizebox{\linewidth}{!}{
    \setlength{\tabcolsep}{3pt}
    \renewcommand{\arraystretch}{1.5}
    \begin{tabular}{r | c c c c c c c  c } \toprule
    \multicolumn{1}{c}{} & \multicolumn{7}{c}{AP} & mAP \\ \cmidrule(lr{.75em}){2-8}
     \multicolumn{1}{c}{Method} & Bus  & Bike & Car & Motor & Person & Rider & Truck & All  \\  \midrule
    FR~\cite{ren2016faster} & 28.1 & 29.7 & 49.7 & 26.3 & 33.2 & 35.5 & 21.5 & 32.0  \\
    % SW \cite{pan2019switchable} & - & - & - & - & - & - & - & - \\
    % IBN-Net \cite{pan2018two} & - & - & - & - & - & - & - & - \\
    % IterNorm \cite{huang2019iterative} & - & - & - & - & - & - & - & - \\
    % ISW \cite{choi2021robustnet} & - & - & - & - & - & - & - & - \\
    S-DGOD \cite{wu2022single} & 32.9 & 28.0 & 48.8 & 29.8 & 32.5 & 38.2 & 24.1 & 33.5 \\
    \hline
    Ours & \textbf{36.1} & \textbf{34.3} & \textbf{58.0} & \textbf{33.1} & \textbf{39.0} & \textbf{43.9} & \textbf{25.1} & \textbf{38.5} \\
    \bottomrule
    \end{tabular}
    }
    \end{center}
    \caption{\textbf{Per-class results on Daytime Clear to Day Foggy.} Our method consistently performs better on all categories for the difficult foggy domain. This shows that CLIP initialization and our semantic augmentations improve the detector's generalizability.
    } 
    \label{tab:class_day_foggy}
\end{table}

%% file: tables/class_results_dusk_rainy.tex
\begin{table}[t]
 
    \begin{center}
    \resizebox{\linewidth}{!}{
    \setlength{\tabcolsep}{3pt}
    \renewcommand{\arraystretch}{1.5}
    \begin{tabular}{r | c c c c c c c  c } \toprule
    \multicolumn{1}{c}{} & \multicolumn{7}{c}{AP} & mAP \\ \cmidrule(lr{.75em}){2-8}
     \multicolumn{1}{c}{Method} & Bus  & Bike & Car & Motor & Person & Rider & Truck & All  \\  \midrule
    FR~\cite{ren2016faster} & 28.5 & 20.3 & 58.2 & 6.5 & 23.4 & 11.3 & 33.9 & 26.0  \\
    % SW \cite{pan2019switchable} & - & - & - & - & - & - & - & - \\
    % IBN-Net \cite{pan2018two} & - & - & - & - & - & - & - & - \\
    % IterNorm \cite{huang2019iterative} & - & - & - & - & - & - & - & - \\
    % ISW \cite{choi2021robustnet} & - & - & - & - & - & - & - & - \\
    S-DGOD \cite{wu2022single} & 37.1 & 19.6 & 50.9 & 13.4 & 19.7 & 16.3 & 40.7 & 28.2 \\
    \hline
    Ours & \textbf{37.8} & \textbf{22.8} & \textbf{60.7} & \textbf{16.8} & \textbf{26.8} & \textbf{18.7} & \textbf{42.4} & \textbf{32.3} \\
    \bottomrule
    \end{tabular}
    }
    \end{center}
    \caption{\textbf{Per-class results on Daytime Clear to Dusk Rainy.} Our approach generalizes to rainy road conditions along with the low light conditions of the dusk hours. The \emph{car} category sees the biggest improvement, but we nonetheless also boost the performance of all the other classes.
    } 
    \label{tab:class_dusk_rainy}
\end{table}

%% file: tables/class_results_night_clear.tex
\begin{table}[h]
 
    \begin{center}
    \resizebox{\linewidth}{!}{
    \setlength{\tabcolsep}{3pt}
    \renewcommand{\arraystretch}{1.5}
    \begin{tabular}{r | c c c c c c c  c } \toprule
    \multicolumn{1}{c}{} & \multicolumn{7}{c}{AP} & mAP \\ \cmidrule(lr{.75em}){2-8}
     \multicolumn{1}{c}{Method} & Bus  & Bike & Car & Motor & Person & Rider & Truck & All  \\ \midrule
    FR~\cite{ren2016faster} & 34.7 & 32.0 & 56.6 & 13.6 & 37.4 & 27.6 & 38.6 & 34.4  \\
    % SW \cite{pan2019switchable} & - & - & - & - & - & - & - & - \\
    % IBN-Net \cite{pan2018two} & - & - & - & - & - & - & - & - \\
    % IterNorm \cite{huang2019iterative} & - & - & - & - & - & - & - & - \\
    % ISW \cite{choi2021robustnet} & - & - & - & - & - & - & - & - \\
    S-DGOD \cite{wu2022single} & \textbf{40.6} & \textbf{35.1} & 50.7 & \textbf{19.7} & 34.7 & \textbf{32.1} & \textbf{43.4} & 36.6 \\
    \hline
    Ours & 37.7 & 34.3 & \textbf{58.0} & 19.2 & \textbf{37.6} & 28.5 & 42.9 & \textbf{36.9} \\
    \bottomrule
    \end{tabular}
    }
    \end{center}
    \caption{\textbf{Per-class results on Daytime Clear to Night Clear.} While being comparable to S-DGOD on most of the categories, we improve on \emph{car} and \emph{person}. 
    } 
    \label{tab:class_night_clear}
\end{table}

%% file: tables/class_results_night_rainy.tex
\begin{table}[t]
 
    \begin{center}
    \resizebox{\linewidth}{!}{
    \setlength{\tabcolsep}{3pt}
    \renewcommand{\arraystretch}{1.5}
    \begin{tabular}{r | c c c c c c c  c } \toprule
    \multicolumn{1}{c}{} & \multicolumn{7}{c}{AP} & mAP \\ \cmidrule(lr{.75em}){2-8}
     \multicolumn{1}{c}{Method} & Bus  & Bike & Car & Motor & Person & Rider & Truck & All  \\  \midrule
    FR~\cite{ren2016faster} & 16.8 & 6.9 & 26.3& 0.6 & 11.6 & 9.4 & 15.4& 12.4  \\
    % SW \cite{pan2019switchable} & - & - & - & - & - & - & - & - \\
    % IBN-Net \cite{pan2018two} & - & - & - & - & - & - & - & - \\
    % IterNorm \cite{huang2019iterative} & - & - & - & - & - & - & - & - \\
    % ISW \cite{choi2021robustnet} & - & - & - & - & - & - & - & - \\
    S-DGOD \cite{wu2022single} & 24.4 & 11.6 & 29.5 & \textbf{9.8} & 10.5 & \textbf{11.4} & 19.2 & 16.6 \\
    \hline
    Ours & \textbf{28.6} & \textbf{12.1} & \textbf{36.1} & 9.2 & \textbf{12.3} & 9.6 & \textbf{22.9} & \textbf{18.7} \\
    \bottomrule
    \end{tabular}
    }
    \end{center}
    \caption{\textbf{Per-class results on Daytime Clear to Night Rainy.} This dataset presents the most challenging scenario, where the low light and rainy conditions obscure the objects. We still perform better than the baseline on most of the categories.
    } 
    \label{tab:class_night_rainy}
\end{table}

%% file: tables/ablation.tex
\begin{table*}[t]
    \begin{center}
    % \resizebox{\linewidth}{!}{
    \renewcommand{\arraystretch}{1.6}
    \vspace{0.5cm}
    \begin{tabular}{c c c c | c c c c c} \toprule
    \multicolumn{4}{c}{\multirow{2}{*}{Model Component}} &  \multicolumn{5}{c}{mAP} \\ \cmidrule(lr{.75em}){5-9}
    \multicolumn{4}{c}{} & \multicolumn{1}{c}{Source} &  \multicolumn{4}{c}{Target} \\ \midrule
    CLIP init &$\mathcal{L}_{ clip\mhyphen t}$ & Attn. Pool  & Sem. Aug  & \makecell{Day \\  Clear}  & \makecell{Night \\ Clear} & \makecell{Dusk \\ Rainy} & \makecell{Night \\ Rainy} & \makecell{Day \\ Foggy} \\ \midrule
     &  &  & & 48.1 & 34.4 & 26.0 & 12.4 & 32.0 \\
    \checkmark &  &  & & 51.2 & \textbf{37.0} & 31.0 & 15.7 & 37.5 \\
    \checkmark & \checkmark  &  & & 50.7 & 36.0 & 31.3 & 16.3 & 36.9\\
    \checkmark & \checkmark  & \checkmark  & & 51.0 & 35.9 & 31.3 & 16.7 & 37.7\\
    \checkmark & \checkmark  & \checkmark  & \checkmark & 51.3 & 36.9 & \textbf{32.3} & \textbf{18.7} & \textbf{38.5} \\
    % \hline
    %  FR\cite{ren2016faster}&  &  & &  48.1 & 34.4 & 26.0 & 12.4 & 32.0 \\
    %  S-DGOD \cite{wu2022single} &  &  & &  \textbf{56.1} & 36.6 & 28.2 & 16.6 & 33.5 \\
    \bottomrule
    \end{tabular}
    % }
    \end{center}
    \caption{\textbf{Ablation study.} We study the influence of five different components of our approach: the backbone weight initialization strategy, the classification loss, the attention pooling, and the semantic augmentation. When those five components are removed (first row of the table) the model is equivalent to the standard FasterRCNN. Initializing the detector with CLIP weights (second row) largely improves the generalization performance; on its own it already outperforms Single-DGOD (penultimate row of \cref{tab:sota}) on most of the datasets, hence suggesting that CLIP has better generalizability than ImageNet pre-trained weights. Combining this with the text embedding-based loss $\mathcal{L}_{clip\mhyphen t}$ (third row) improves the results on the challenging scenarios of dusk rainy and night rainy, but has a detrimental effect for the other weather conditions. Adding attention pooling to the architecture (fourth row) helps to mitigate these detrimental effects as it brings the visual features closer to the joint embedding space. Finally, the best results are obtained when the semantic augmentation is added (last row), greatly helping with adverse weather, rainy and foggy, scenarios.
    } 
    \label{tab:abla}
    \end{table*}
    

%% file: tables/different_aug.tex
\begin{table}[t]
 \setlength{\tabcolsep}{4pt}
    \begin{center}
    \renewcommand{\arraystretch}{1.6}
    \begin{tabular}{c | c | c c c c } \toprule
    \multicolumn{1}{c}{} & \multicolumn{5}{c}{mAP} \\ \cmidrule(lr{.75em}){2-6}
    %  \multicolumn{1}{c}{} &  &  \multicolumn{4}{c}{Out-Of-Domain} \\ 
     \multicolumn{1}{c}{Aug. Type} & \makecell{Day \\  Clear}  & \makecell{Night \\ Clear} & \makecell{Dusk \\ Rainy} & \makecell{Night \\ Rainy} & \makecell{Day \\ Foggy} \\ \midrule
    no-aug. & 51.0 & 35.9 & 31.3 & 16.7 & 37.7 \\
    random & 51.2 & 36.0 & 30.4 & 15.3 & 37.3 \\
    clip-random  & 51.5 & 36.4 & 30.2 & 15.9 & 37.9 \\
    
    \hline
    Ours w/ seg.aug & 51.3 & \textbf{36.9} & \textbf{32.3} & \textbf{18.7} & \textbf{38.5} \\
    \bottomrule
    \end{tabular}
    % }
    \end{center}
    \caption{\textbf{Semantic Augmentation.} Our semantic augmentation consistently outperforms other augmentation strategies. While \emph{random} augmentations are worse than \emph{no-aug.}, \emph{clip-random} is comparable to \emph{no-aug.}. Only when we give relevant prompts, there is a consistent improvement across datasets. 
    } 
    \label{tab:seg-aug}
    \end{table}

%% file: tex/limit_societal_impact.tex
\section{Limitations}
Our method augments visual features using textual prompts. To generate these prompts, it is assumed that some information about the domain gap is known. In our experiments, we assumed that the domain gap was due to changes in weather and daytime conditions. In practice, we only used the word \emph{weather} and \emph{time of the day} to derive all the prompts used in our augmentation; nonetheless, some extra information was used. In most applications, however, the domain gap can be known in advance, and providing a few keywords characterizing it shouldn't be an issue. In the rare cases where no information can be known, our approach still has the potential to be used by using multiple broad concept keywords such as weather, ambiance, or location.

%% file: tex/conclusion.tex
\section{Conclusion}
We have proposed an approach to improving the generalization of object detectors on \emph{unseen} target domains. Our approach fundamentally departs from existing method by leveraging a pre-trained vision-language model, CLIP, to help the detector to generalize. Specifically, we have exploited textual prompts to develop a semantic augmentation strategy that alters image embeddings so that they reflect potential target domains, and to design a text-based image classifier. We have shown that our approach outperforms the state of the art on four adverse-weather target datasets. In future work, we plan to extend our approach to learning the prompts to further improve generalization. 

\paragraph{Acknowledgment:}
 This work was funded in part by the Swiss National Science Foundation and the Swiss Innovation Agency (Innosuisse) via the BRIDGE Discovery grant 40B2-0 194729.